\documentclass[letterpaper, 10 pt, conference]{ieeeconf}  

\IEEEoverridecommandlockouts                              

\usepackage{amsmath} 
\usepackage{graphicx} 
\usepackage{amssymb}  
\usepackage{algorithm}
\usepackage{algorithmic}
\usepackage{multirow}
\usepackage{booktabs}

\title{\LARGE \bf
Statistical Uncertainty Learning\\ for Robust Visual-Inertial State Estimation
}

\author{Seungwon Choi, Dong-Gyu Park, Seo-Yeon Hwang\thanks{S. Choi, D. Park, and S.-Y. Hwang are with the Research Center, Clobot Inc., Seongnam, Republic of Korea. (e-mail: eugene@clobot.co.kr, derrick@clobot.co.kr, ian@clobot.co.kr)}, Tae-Wan Kim\thanks{T.-W. Kim is with the Department of Naval Architecture and Ocean Engineering, Seoul National University (SNU), Seoul, Republic of Korea (e-mail: taewan@snu.ac.kr)}}

\begin{document}

\maketitle
\thispagestyle{empty}
\pagestyle{empty}

\begin{abstract}
A fundamental challenge in robust visual-inertial odometry (VIO) is to dynamically assess the reliability of sensor measurements. This assessment is crucial for properly weighting the contribution of each measurement to the state estimate. Conventional methods often simplify this by assuming a static, uniform uncertainty for all measurements. This heuristic, however, may be limited in its ability to capture the dynamic error characteristics inherent in real-world data. To improve this limitation, we present a statistical framework that learns measurement reliability assessment online, directly from sensor data and optimization results. Our approach leverages multi-view geometric consistency as a form of self-supervision. This enables the system to infer landmark uncertainty and adaptively weight visual measurements during optimization. We evaluated our method on the public EuRoC dataset, demonstrating improvements in tracking accuracy with average reductions of approximately 24\% in translation error and 42\% in rotation error compared to baseline methods with fixed uncertainty parameters. The resulting framework operates in real time while showing enhanced accuracy and robustness. To facilitate reproducibility and encourage further research, the source code will be made publicly available.
\end{abstract}


\section{INTRODUCTION}

State estimation is a fundamental research topic in robotics, given its indispensable role in the navigation of autonomous systems\cite{thrun2005probabilistic}. Among the various sensor modalities, the fusion of visual and inertial measurements has become a popular approach for state estimation \cite{lupton2012visual, scaramuzza2011visual}, primarily due to the complementary nature of cameras and Inertial Measurement Units (IMUs). While cameras provide rich geometric information about the environment, IMUs offer high-frequency motion measurements that are robust to visual degradation. This combination allows for robust tracking in a wide range of environments, from well-lit indoor spaces to challenging outdoor scenarios \cite{sun2018robust}.

The problem is typically formulated as a non-linear least squares (NLS) optimization within a sliding window framework \cite{leutenegger2015keyframe}, where camera poses, 3D landmarks, and sensor biases are jointly estimated. Modern visual-inertial odometry (VIO) systems often employ factor graph optimization \cite{dellaert2017factor} to fuse heterogeneous sensor measurements, with each observation weighted by its assumed uncertainty. However, one potential limitation in many approaches is the reliance on static noise models, which may not fully capture the dynamic nature of measurement uncertainty \cite{strasdat2012visual}.

\begin{figure}
    \centering
    \includegraphics[width=1.0\linewidth]{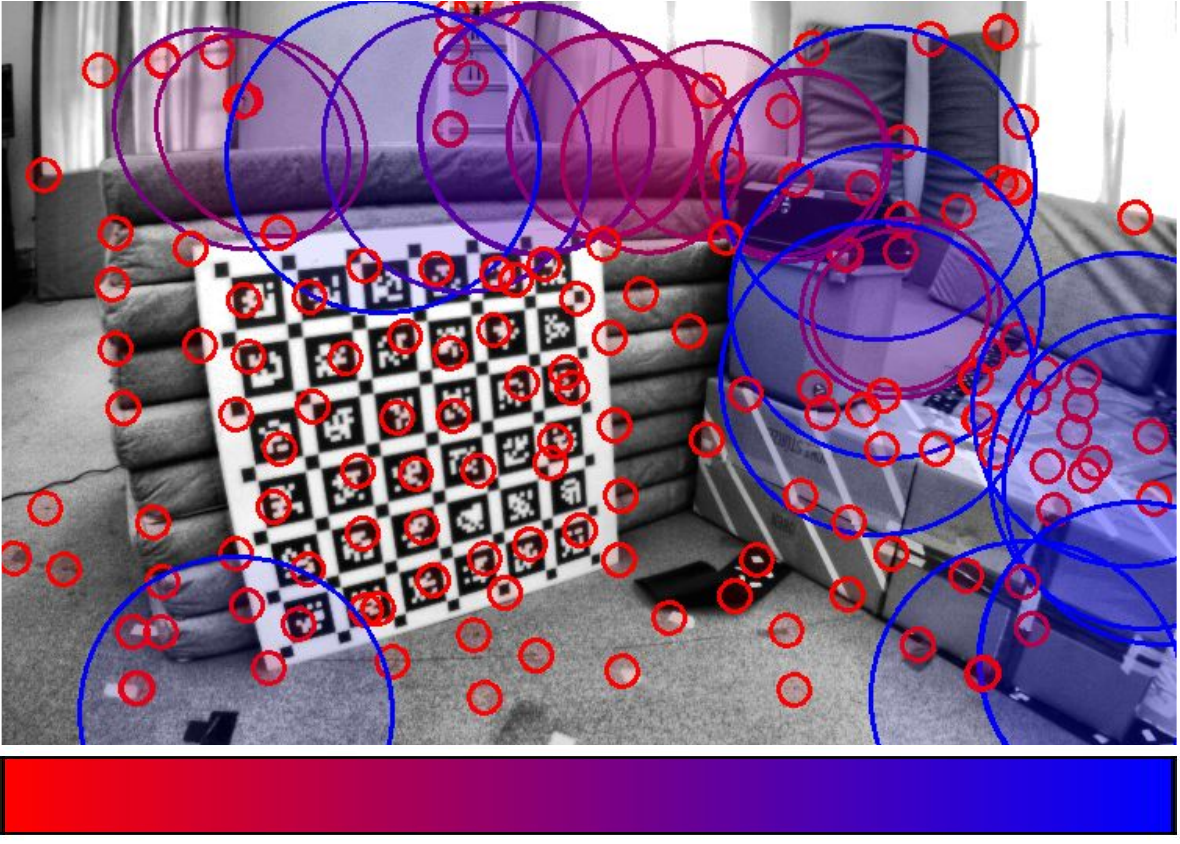}
    \caption{Visualization of learned uncertainty estimates for visual landmarks. The ellipses represent the projected 2D covariance of 3D landmark uncertainties onto the image plane, illustrating the spatial distribution of measurement reliability. Color coding indicates uncertainty magnitude: red ellipses correspond to lower uncertainties (higher reliability), while blue ellipses indicate higher uncertainties (lower reliability). This adaptive uncertainty modeling enables observation-specific weighting in bundle adjustment optimization.}
    \label{fig:uncertainty_example}
\end{figure}

\begin{figure*}
    \centering
    \includegraphics[width=1.0\linewidth]{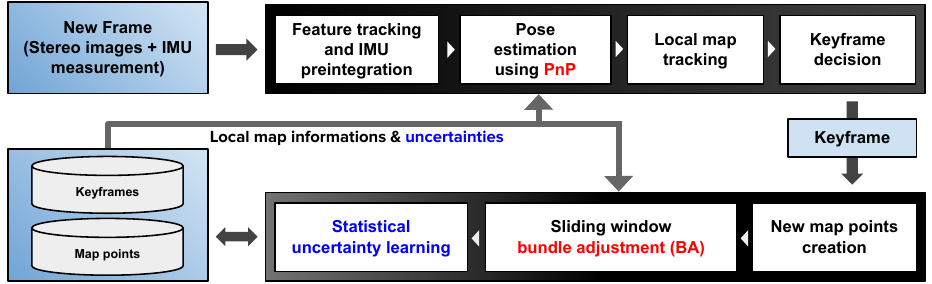}
    \caption{Overview of the proposed visual-inertial system architecture featuring a dual-pipeline design. The upper pipeline handles real-time pose tracking via Perspective-n-Point (PnP) optimization, while the lower pipeline performs map refinement through sliding window bundle adjustment (BA) integrated with our statistical uncertainty learning module. The learned uncertainty estimates are propagated back to the tracking pipeline, establishing a feedback mechanism that enhances system performance.}
    \label{fig:system_overview}
\end{figure*}

Consider the inherent variability in visual measurements: a feature tracked in well-textured, well-lit regions exhibits different error characteristics compared to one tracked near occlusion boundaries or in poorly illuminated areas. Similarly, the quality of stereo depth estimates can vary with baseline geometry and scene depth \cite{hartley2003multiple}. Traditional VIO frameworks often address this heterogeneity through uniform information matrices. This treatment of all visual observations as equally reliable is a simplifying assumption that may not always account for the rich statistical structure inherent in the measurements.

The effects of this limitation can become particularly pronounced in challenging scenarios. When a system encounters rapid motion, varying lighting conditions, or textureless environments, a mismatch between the assumed and actual measurement uncertainties can contribute to less accurate estimates, convergence issues, or tracking failures \cite{schneider2020maplab}. As autonomous systems operate in increasingly diverse environments, the ability to adapt measurement uncertainty models online could be crucial for maintaining consistent performance.

To explore this area, this work proposes a \textbf{statistical uncertainty learning} approach for visual-inertial odometry. Our framework is designed as an alternative to static noise models, aiming to statistically learn and propagate measurement uncertainties throughout the estimation pipeline. By leveraging multi-view geometric consistency and covariance analysis, we investigate whether a system can better estimate the error characteristics of its sensor measurements. Figure~\ref{fig:uncertainty_example} illustrates our approach to modeling measurement variability through adaptive covariance representations. 

Our system employs a dual-pipeline architecture that integrates statistical uncertainty learning with real-time VIO estimation. The upper pipeline handles real-time pose tracking via Perspective-n-Point (PnP) optimization, while the lower pipeline performs map refinement through sliding window bundle adjustment integrated with our statistical uncertainty learning module. The learned uncertainty estimates are propagated back to the tracking pipeline, establishing a feedback mechanism that enhances system performance. Figure~\ref{fig:system_overview} provides an overview of this dual-pipeline architecture.

Specifically, the key contributions of our proposed framework include:

\begin{itemize}
\item \textbf{Statistical multi-view uncertainty learning}: We propose a method to estimate 3D landmark uncertainty by analyzing the distribution of triangulated positions across multiple viewpoints. This approach aims to provide a data-driven measure of geometric consistency and measurement reliability.

\item \textbf{Adaptive statistical weighting}: We investigate transforming learned 3D world uncertainties to 2D pixel-space information matrices through uncertainty propagation. This enables observation-specific weighting that may better reflect the statistical reliability of each measurement.

\item \textbf{Open-source statistical VIO framework}: We present a complete, open-source implementation of our approach, which integrates with existing optimization frameworks while maintaining computational efficiency suitable for real-time applications.
\end{itemize}

Through evaluation on the EuRoC dataset \cite{burri2016euroc}, we demonstrate that our statistical learning approach achieves improved accuracy and robustness compared to baseline methods using static noise models across various conditions. The proposed method represents a step toward more adaptive and statistically-grounded visual-inertial state estimation.

\section{RELATED WORK}

Visual-inertial odometry (VIO) has become a cornerstone of autonomous robot navigation and state estimation \cite{qin2018vins, campos2021orb}, with state-of-the-art systems broadly categorized into filtering-based and optimization-based methods \cite{strasdat2012visual}. A critical component within these systems is the modeling of measurement uncertainty, which directly influences the weighting of visual observations in bundle adjustment and ultimately determines the accuracy and robustness of the state estimate. This section reviews how prominent VIO frameworks handle visual measurement uncertainty and highlights the limitations of current static approaches.

Early influential filtering-based methods such as MSCKF \cite{mourikis2007multi} established the foundation for tightly-coupled visual-inertial fusion. In MSCKF, the covariance of visual feature measurements is typically assumed to be a fixed identity matrix, implying that all feature observations are treated with equal confidence regardless of their actual reliability. This simplification, while computationally efficient, overlooks significant variations in feature quality, lighting conditions, and tracking ambiguity that naturally occur in real-world scenarios. Subsequent filtering-based systems like ROVIO \cite{bloesch2015robust} and OpenVINS \cite{geneva2020openvins} perpetuate this limitation by employing constant and uniform noise models for image feature re-projection errors. These models are typically determined through empirical tuning on specific datasets, making them poorly suited for diverse operating conditions encountered in practical applications.

Optimization-based approaches, which have gained prominence due to their superior handling of nonlinear constraints \cite{kummerle2011g2o}, often retain similarly simplified uncertainty models. VINS-Mono \cite{qin2018vins}, one of the most widely adopted VIO systems, models the re-projection error of visual features using unit covariance matrices in its bundle adjustment formulation. Specifically, the information matrix for each feature observation is set to $\boldsymbol{\Omega} = \mathbf{I}$, effectively assigning uniform weight to every measurement regardless of factors such as illumination quality, feature distinctiveness, or geometric configuration. Other notable systems such as direct-methods \cite{engel2014lsd, engel2018direct} and semi-direct methods \cite{forster2014svo} also typically rely on simplified photometric error models.

Some systems have made modest progress toward adaptive uncertainty modeling. The ORB-SLAM family \cite{mur2015orb, 7946260, campos2021orb} introduces scale-dependent weighting based on the pyramid level at which features are detected. The information matrix is scaled by $\sigma^{-2}$, where $\sigma$ represents the scale factor of the detection level. While this approach acknowledges that features detected at different scales have varying precision, it remains a predefined heuristic rather than a learned model. OKVIS \cite{leutenegger2015keyframe} provides infrastructure for more sophisticated uncertainty modeling, with provisions for incorporating feature tracking scores and detection quality metrics. However, in practice, many implementations often rely on simple isotropic uncertainty models, partly due to the challenges in developing robust methods for learning appropriate weightings online.

The predominant reliance on static, hand-tuned noise models represents a fundamental limitation in current VIO systems. These approaches fail to capture the inherent variability in measurement quality that arises from environmental factors such as lighting conditions, weather, and scene texture; geometric factors including viewing angle, distance, and baseline geometry; and dynamic factors like camera motion, exposure settings, and focus changes. The mismatch between assumed and actual measurement uncertainties can lead to overconfident estimates, poor convergence in challenging scenarios, and suboptimal information weighting during optimization.

Recent work in uncertainty quantification has explored learning-based approaches in related domains. In structure from motion, some methods have investigated uncertainty propagation through multi-view geometry \cite{forstner2016photogrammetric}, while robust estimation techniques have focused on outlier detection and covariance inflation \cite{huber2009robust}. However, these approaches either assume known uncertainty models or focus on outlier rejection rather than systematic uncertainty learning. The integration of uncertainty learning directly into the VIO optimization loop—where bundle adjustment results provide statistical evidence for refining uncertainty models—remains relatively underexplored, although some works have begun to explore this direction \cite{usenko2019visual}. Deep learning methods have also been proposed to estimate uncertainty directly from images \cite{kendall2017uncertainties}, but often require large amounts of training data.

With the goal of improving upon the limitations of static noise models, this work introduces a statistical uncertainty learning framework. Our method learns measurement uncertainties from multi-view geometric consistency by using bundle adjustment results as statistical evidence. This approach allows uncertainty estimates to adapt to observed data characteristics, aiming for more effective information weighting in visual-inertial optimization. The core of our contribution is the integration of this uncertainty learning within the bundle adjustment loop, where optimized poses and landmarks provide the empirical evidence to refine uncertainty models for subsequent optimizations. This represents a step toward more adaptive, data-driven uncertainty estimation.

\section{VISUAL-INERTIAL SYSTEM OVERVIEW}

Our approach builds upon established visual-inertial odometry principles, employing a sliding window-based optimization strategy that combines real-time pose tracking with batch refinement.

\subsection{System Architecture}

The visual-inertial system follows a two-stage estimation pipeline: (1) real-time pose estimation using Perspective-n-Point (PnP) optimization for immediate tracking \cite{lepetit2009epnp}, and (2) sliding window bundle adjustment for refined state estimation \cite{leutenegger2015keyframe}. This architecture balances computational efficiency with estimation accuracy, making it suitable for real-time applications while maintaining high precision through batch optimization. The final optimization is performed using a modern nonlinear solver like Ceres solver.

\subsection{State Representation}

The system state at time $k$ consists of the camera pose, velocity, and IMU biases:
\begin{equation}
\mathbf{x}_k = \left[ \mathbf{T}_{wb,k}, \mathbf{v}_{w,k}, \mathbf{b}_{a,k}, \mathbf{b}_{g,k} \right]^T
\end{equation}
\noindent where $\mathbf{T}_{wb,k} \in SE(3)$ represents the transformation from body frame to world frame, $\mathbf{v}_{w,k} \in \mathbb{R}^3$ is the velocity in world coordinates, and $\mathbf{b}_{a,k}, \mathbf{b}_{g,k} \in \mathbb{R}^3$ are the accelerometer and gyroscope biases, respectively.

For optimization purposes, the $SE(3)$ pose is parameterized using the Lie algebra representation \cite{barfoot2017state}:
\begin{equation}
\mathbf{T}_{wb} = \exp(\boldsymbol{\xi}^\wedge), \quad \boldsymbol{\xi} = [\boldsymbol{\rho}^T, \boldsymbol{\phi}^T]^T \in \mathbb{R}^6
\end{equation}
\noindent where $\boldsymbol{\rho} \in \mathbb{R}^3$ and $\boldsymbol{\phi} \in \mathbb{R}^3$ represent the translation and rotation components in the tangent space.

\subsection{Real-time Pose Tracking via PnP}

For each incoming frame, we perform real-time pose estimation using the Perspective-n-Point algorithm \cite{lepetit2009epnp}. Given a set of 3D landmarks $\{\mathbf{p}_j\}$ and their corresponding 2D observations $\{\mathbf{z}_j\}$, the PnP optimization minimizes:
\begin{equation}
\boldsymbol{\xi}^* = \arg\min_{\boldsymbol{\xi}} \sum_{j} \boldsymbol{e}_j^T \boldsymbol{\Omega}_j \boldsymbol{e}_j
\end{equation}
\noindent where the re-projection error is defined as:
\begin{equation}
\boldsymbol{e}_j = \mathbf{z}_j - \pi(\mathbf{T}_{cb} \exp(\boldsymbol{\xi}^\wedge) \mathbf{p}_j)
\end{equation}
and $\pi(\cdot)$ represents the camera projection function, $\mathbf{T}_{cb}$ is the body-to-camera transformation, and $\boldsymbol{\Omega}_j$ is the information matrix for observation $j$.

The projection function for a pinhole camera model is:
\begin{equation}
\pi(\mathbf{p}_c) = \begin{bmatrix} f_x \frac{X_c}{Z_c} + c_x \\ f_y \frac{Y_c}{Z_c} + c_y \end{bmatrix}
\end{equation}
\noindent where $\mathbf{p}_c = [X_c, Y_c, Z_c]^T$ is the 3D point in camera coordinates, and $(f_x, f_y, c_x, c_y)$ are the camera intrinsic parameters.

\subsection{IMU Preintegration}

Between consecutive keyframes, IMU measurements are preintegrated to provide motion constraints, following the on-manifold preintegration theory \cite{forster2017manifold}. Given IMU measurements $\{\boldsymbol{\omega}_t, \mathbf{a}_t\}$ over the time interval $[t_i, t_{i+1}]$, we compute the preintegrated quantities:
\begin{align}
\boldsymbol{\alpha}_{i,i+1} &= \iint_{t_i}^{t_{i+1}} \mathbf{R}_t (\mathbf{a}_t - \mathbf{b}_{a,t} - \mathbf{n}_a) dt^2 \\
\boldsymbol{\beta}_{i,i+1} &= \int_{t_i}^{t_{i+1}} \mathbf{R}_t (\mathbf{a}_t - \mathbf{b}_{a,t} - \mathbf{n}_a) dt \\
\boldsymbol{\gamma}_{i,i+1} &= \int_{t_i}^{t_{i+1}} \frac{1}{2}(\boldsymbol{\omega}_t - \mathbf{b}_{g,t} - \mathbf{n}_g)^\wedge dt
\end{align}
\noindent where $\mathbf{R}_t$ is the rotation at time $t$, and $\mathbf{n}_a, \mathbf{n}_g$ represent IMU noise terms.

The IMU constraint between two consecutive states is formulated as:
\begin{equation}
\boldsymbol{r}_{IMU} = \begin{bmatrix}
\mathbf{R}_i^T(\mathbf{p}_{i+1} - \mathbf{p}_i - \mathbf{v}_i \Delta t - \frac{1}{2}\mathbf{g}\Delta t^2) - \boldsymbol{\alpha}_{i,i+1} \\
\mathbf{R}_i^T(\mathbf{v}_{i+1} - \mathbf{v}_i - \mathbf{g}\Delta t) - \boldsymbol{\beta}_{i,i+1} \\
\log((\boldsymbol{\gamma}_{i,i+1})^{-1} \mathbf{R}_i^T \mathbf{R}_{i+1})^\vee
\end{bmatrix}
\end{equation}

\begin{figure}
    \centering
    \includegraphics[width=1.0\linewidth]{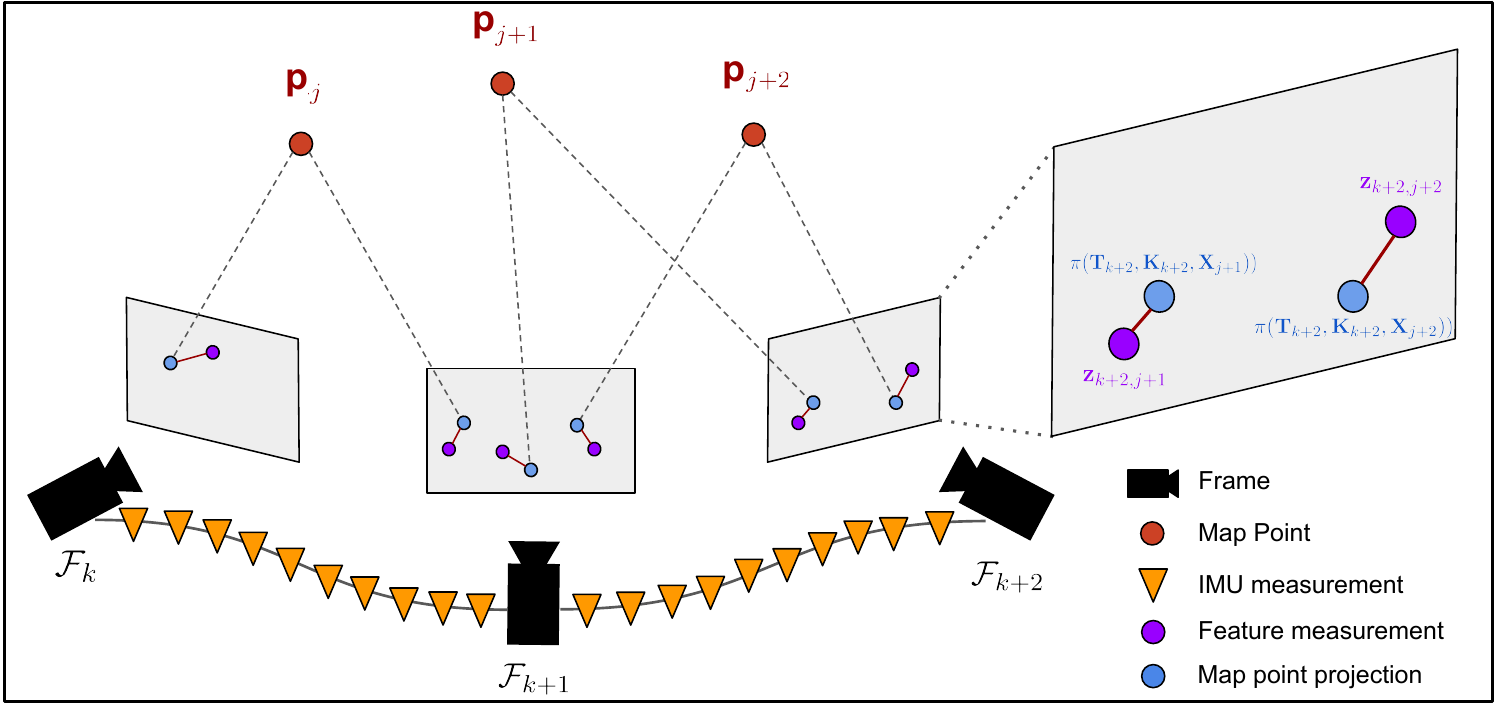}
    \caption{Sliding window bundle adjustment framework. The system maintains a fixed-size window of keyframes while marginalizing older frames. Visual observations connect 3D landmarks to multiple keyframes, while IMU measurements provide constraints between consecutive poses. The optimization jointly estimates camera poses, landmark positions, and IMU biases within the sliding window.}
    \label{fig:sliding_window_ba}
\end{figure}

\subsection{Sliding Window Bundle Adjustment}

The sliding window optimization jointly refines camera poses, 3D landmarks, velocities, and IMU biases over a fixed-size window of keyframes \cite{leutenegger2015keyframe}. The optimization objective combines visual re-projection errors and IMU constraints:
\begin{equation}
\min_{\mathcal{X}, \mathcal{L}} \sum_{(i,j) \in \mathcal{O}} \boldsymbol{e}_{ij}^T \boldsymbol{\Omega}_{ij} \boldsymbol{e}_{ij} + \sum_{k=0}^{N-1} \boldsymbol{r}_{IMU,k}^T \boldsymbol{\Sigma}_{IMU,k}^{-1} \boldsymbol{r}_{IMU,k}
\end{equation}
\noindent where $\mathcal{X} = \{\mathbf{x}_0, \ldots, \mathbf{x}_N\}$ represents the sliding window states, $\mathcal{L} = \{\mathbf{p}_1, \ldots, \mathbf{p}_M\}$ are the 3D landmarks, $\mathcal{O}$ denotes the set of visual observations, and $\boldsymbol{\Sigma}_{IMU,k}$ is the IMU measurement covariance. This optimization framework is illustrated in Figure~\ref{fig:sliding_window_ba}.

\section{STATISTICAL UNCERTAINTY LEARNING}

Our statistical uncertainty learning framework extends the traditional VIO pipeline by creating a closed-loop system where bundle adjustment optimization results continuously refine uncertainty models, which in turn improve subsequent optimizations. This section details our adaptive approach that learns from optimization outcomes to enhance measurement reliability estimation.

\subsection{Learning-based Uncertainty Framework Overview}

Unlike traditional VIO systems that rely on fixed noise models, our approach implements a statistical learning cycle where bundle adjustment optimization provides empirical evidence for uncertainty refinement, which then improves adaptive weighting in subsequent optimizations, creating a self-improving system. This iterative process ensures that uncertainty estimates continuously adapt to the observed statistical properties of the sensor measurements and environmental conditions.

\subsection{Uncertainty Propagation for Adaptive Weighting}

For adaptive information matrix computation in bundle adjustment, we propagate learned 3D world uncertainties to 2D pixel space through a series of transformations. Given a 3D point $\mathbf{p}_{world}$ with learned covariance $\boldsymbol{\Sigma}_{world}$, the geometric transformation chain follows:

\begin{align}
\mathbf{p}_{world} \xrightarrow{\mathbf{T}_{cw}} \mathbf{p}_{cam} \xrightarrow{\pi} \mathbf{z}
\end{align}

To propagate uncertainties through this transformation, we first compute the projection Jacobian with respect to the camera coordinates:
\begin{align}
\mathbf{J}_{\pi} = \frac{\partial \pi(\mathbf{p}_{cam})}{\partial \mathbf{p}_{cam}} = \begin{bmatrix}
\frac{f_x}{Z_c} & 0 & -\frac{f_x X_c}{Z_c^2} \\
0 & \frac{f_y}{Z_c} & -\frac{f_y Y_c}{Z_c^2}
\end{bmatrix}
\end{align}

Applying the uncertainty propagation formula through both the rotation and projection transformations yields the pixel-space covariance:
\begin{align}
\boldsymbol{\Sigma}_{pixel} = \mathbf{J}_{\pi} \mathbf{R}_{cw} \boldsymbol{\Sigma}_{world} \mathbf{R}_{cw}^T \mathbf{J}_{\pi}^T
\end{align}

To ensure numerical stability during optimization, we construct the adaptive information matrix by inverting the pixel covariance and adding a small regularization term:
\begin{align}
\boldsymbol{\Omega}_{adaptive} = \boldsymbol{\Sigma}_{pixel}^{-1} + \lambda \mathbf{I}
\end{align}
where $\lambda$ is a small regularization parameter that prevents singular matrices during inversion.

This adaptively weighted information matrix is then incorporated into the bundle adjustment optimization objective, allowing each observation to contribute according to its learned reliability:
\begin{align}
\min_{\mathcal{X}, \mathcal{L}} \sum_{(i,j) \in \mathcal{O}} \boldsymbol{e}_{ij}^T \boldsymbol{\Omega}_{adaptive,ij} \boldsymbol{e}_{ij}
\end{align}
\subsection{Post-Optimization Statistical Learning}

After each bundle adjustment iteration, we extract statistical evidence from the optimized results to refine our uncertainty estimates. The core idea is to leverage the multi-view geometric consistency of optimized landmarks to assess measurement reliability. For each map point, we collect triangulated positions using the newly optimized camera poses and landmark positions, as detailed in Algorithm 1.

\begin{algorithm}[h]
\caption{Statistical Learning from Bundle Adjustment Results}
\begin{algorithmic}[1]
\STATE \textbf{Input:} Optimized poses $\{\mathbf{T}_{wb,i}^{opt}\}$, optimized points $\{\mathbf{p}_j^{opt}\}$
\FOR{each MapPoint $\mathbf{p}_j^{opt}$ in optimization window}
    \STATE \textbf{Initialize:} Position collection $\mathcal{P}_j = \emptyset$
    \FOR{each observation $(frame_i, feature_k)$ of $\mathbf{p}_j$}
        \STATE \textbf{Update depth:} $d_{ik} = (\mathbf{T}_{cw,i}^{opt} \mathbf{p}_j^{opt}).z()$
        \STATE \textbf{Triangulate:} $\mathbf{p}_{world}^{(ik)} = \mathbf{T}_{wc,i}^{opt} [d_{ik} \mathbf{p}_{norm}, d_{ik}]^T$
        \STATE \textbf{Collect:} $\mathcal{P}_j \leftarrow \mathcal{P}_j \cup \{\mathbf{p}_{world}^{(ik)}\}$
    \ENDFOR
    \STATE \textbf{Compute:} Empirical covariance using optimized position as mean
    \STATE \textbf{Update:} MapPoint uncertainty $\boldsymbol{\Sigma}_j^{new}$
\ENDFOR
\end{algorithmic}
\end{algorithm}
\subsection{Iterative Statistical Refinement}

The complete learning cycle integrates seamlessly into sliding window optimization through an iterative refinement process. Algorithm 2 outlines the detailed procedure for statistical learning-enhanced bundle adjustment, which alternates between forward uncertainty propagation and backward statistical learning phases. This iterative process is designed to exhibit several desirable properties that enhance the overall system performance. First, it promotes self-consistency, where improved uncertainty estimates may contribute to more accurate optimization, potentially providing higher quality statistical evidence for further refinement. Second, the framework demonstrates adaptive convergence, as uncertainty models automatically adjust to specific sensor characteristics and prevailing environmental conditions. Finally, the process exhibits inherent robustness; even when initialized with suboptimal uncertainty estimates, the continuous accumulation of statistical evidence from new observations gradually improves the uncertainty models over time.

\begin{algorithm}[h]
\caption{Statistical Learning-Enhanced Bundle Adjustment}
\begin{algorithmic}[1]
\STATE \textbf{Initialize:} Use prior uncertainty estimates $\{\boldsymbol{\Sigma}_j^{(0)}\}$
\FOR{each bundle adjustment iteration $t$}
    \STATE \textbf{Forward Pass:} Create adaptive information matrices
    \FOR{each observation $(i,j)$}
        \STATE $\boldsymbol{\Omega}_{ij}^{(t)} = (\mathbf{J}_{\pi} \mathbf{R}_{cw} \boldsymbol{\Sigma}_j^{(t)} \mathbf{R}_{cw}^T \mathbf{J}_{\pi}^T)^{-1}$
    \ENDFOR
    \STATE \textbf{Optimization:} Solve bundle adjustment with adaptive weighting
    \STATE $\{\mathbf{T}_{wb,i}^{opt}, \mathbf{p}_j^{opt}\} = \arg\min \sum_{(i,j)} \boldsymbol{e}_{ij}^T \boldsymbol{\Omega}_{ij}^{(t)} \boldsymbol{e}_{ij}$
    \STATE \textbf{Backward Pass:} Learn from optimization results
    \FOR{each MapPoint $j$}
        \STATE \textbf{Collect:} Multi-view positions using $\{\mathbf{T}_{wb,i}^{opt}\}$
        \STATE \textbf{Compute:} Empirical covariance around $\mathbf{p}_j^{opt}$
        \STATE \textbf{Update:} $\boldsymbol{\Sigma}_j^{(t+1)} \leftarrow \boldsymbol{\Sigma}_{empirical,j}$
    \ENDFOR
\ENDFOR
\end{algorithmic}
\end{algorithm}

\begin{figure*}
    \centering
    \includegraphics[width=1.0\linewidth]{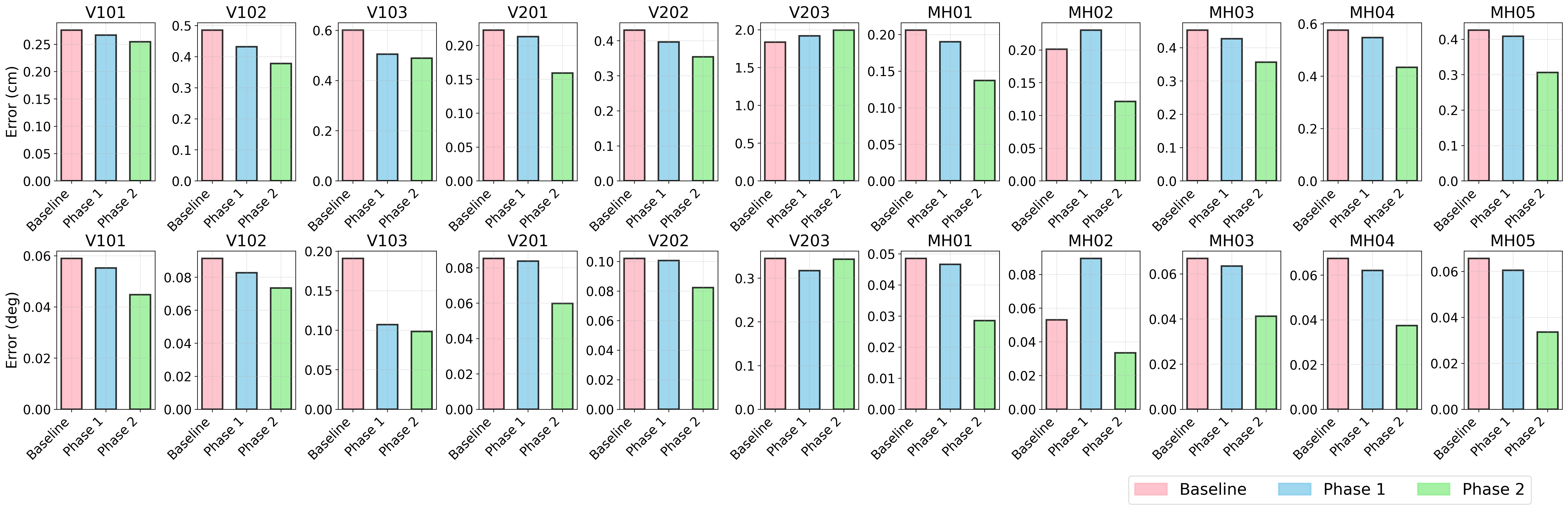}
    \caption{Quantitative evaluation of frame-to-frame relative pose accuracy across EuRoC sequences. Upper panels display translation RMSE (cm), lower panels show rotation RMSE (degrees). Our statistical uncertainty learning approach (Phase 2) demonstrates consistent improvements over baseline and geometric uncertainty methods (Phase 1) across most sequences. The progressive improvement trend (Baseline → Phase 1 → Phase 2) validates the effectiveness of adaptive uncertainty learning for visual-inertial odometry.}
    \label{fig:comparison}
\end{figure*}

Through this iterative refinement process, the learned uncertainties $\boldsymbol{\Sigma}_j^{(t)}$ progressively converge toward representations that better reflect true measurement reliability, creating observation-specific weighting that captures actual geometric consistency across multiple viewpoints.

\subsection{Integration with Visual-Inertial Pipeline}

The statistical learning framework operates as a background process during sliding window optimization, with minimal computational overhead:

\begin{align}
\text{Total Cost} = &\sum_{(i,j) \in \mathcal{O}} \boldsymbol{e}_{visual,ij}^T \boldsymbol{\Omega}_{learned,ij} \boldsymbol{e}_{visual,ij} \nonumber \\
&+ \sum_{k} \boldsymbol{r}_{IMU,k}^T \boldsymbol{\Sigma}_{IMU,k}^{-1} \boldsymbol{r}_{IMU,k}
\end{align}

\noindent where $\boldsymbol{\Omega}_{learned,ij}$ represents the statistically learned information matrices that continuously improve through the optimization feedback loop.

\section{Experimental Results} 
\subsection{Experimental Setup} We evaluated our framework on the EuRoC MAV dataset \cite{burri2016euroc}. For the quantitative analysis, we compared three configurations in a progressive manner. We started with a \textbf{Baseline} method, representing a standard VIO approach that assumes uniform uncertainty for all visual observations. The first step, \textbf{Phase 1}, introduced an enhanced model incorporating triangulation-based uncertainty derived from geometric constraints. This allowed us to investigate our first key hypothesis: whether explicitly modeling uncertainty, even with a simple geometric approach, is fundamentally more effective than treating all measurements as equally reliable. The final step, \textbf{Phase 2}, implemented our full proposed framework, which adds a learning-based module to adaptively calibrate uncertainty from statistical feedback. The goal of this phase was to test our second hypothesis: that a statistical learning approach can yield additional improvements by capturing complex uncertainty patterns that a fixed geometric model may not represent. 
\begin{table*}[t] \centering \caption{Frame-to-frame relative pose RMSE comparison on EuRoC dataset.} \label{tab:euroc_combined} \resizebox{\textwidth}{!}{ \begin{tabular}{llccccccccccc} \toprule \multirow{2}{*}{\textbf{Method}} & \multirow{2}{*}{\textbf{Error Type}} & \multicolumn{11}{c}{\textbf{EuRoC Sequence}} \\ \cmidrule(lr){3-13} & & \textbf{V1\_01} & \textbf{V1\_02} & \textbf{V1\_03} & \textbf{V2\_01} & \textbf{V2\_02} & \textbf{V2\_03} & \textbf{MH\_01} & \textbf{MH\_02} & \textbf{MH\_03} & \textbf{MH\_04} & \textbf{MH\_05} \\ \midrule \multirow{2}{*}{Baseline} & Trans (cm) & 0.276 & 0.485 & 0.601 & 0.222 & 0.430 & \textbf{1.84} & 0.206 & 0.201 & 0.452 & 0.576 & 0.426 \\ & Rot (°) & 0.059 & 0.091 & 0.191 & 0.085 & 0.102 & 0.345 & 0.049 & 0.053 & 0.067 & 0.067 & 0.066 \\ \midrule \multirow{2}{*}{Phase 1} & Trans (cm) & 0.267 & 0.432 & 0.505 & 0.213 & 0.396 & 1.919 & 0.190 & 0.230 & 0.427 & 0.548 & 0.409 \\ & Rot (°) & 0.055 & 0.083 & 0.107 & 0.084 & 0.101 & \textbf{0.317} & 0.047 & 0.089 & 0.064 & 0.062 & 0.061 \\ \midrule \multirow{2}{*}{Phase 2} & Trans (cm) & \textbf{0.255} & \textbf{0.378} & \textbf{0.489} & \textbf{0.159} & \textbf{0.354} & 1.993 & \textbf{0.137} & \textbf{0.121} & \textbf{0.356} & \textbf{0.434} & \textbf{0.306} \\ & Rot (°) & \textbf{0.045} & \textbf{0.074} & \textbf{0.098} & \textbf{0.060} & \textbf{0.082} & 0.344 & \textbf{0.029} & \textbf{0.033} & \textbf{0.041} & \textbf{0.037} & \textbf{0.034} \\ \bottomrule \end{tabular} } \end{table*} 
\subsection{Experimental Environment} 
All experiments were conducted on a desktop system running a Linux-based operating system. The hardware was equipped with an Intel(R) Core(TM) i7-14700KF processor, which features a hybrid architecture of 20 cores (8 Performance-cores and 12 Efficient-cores) and 28 threads, with a maximum clock speed of 5.6 GHz and 20 MiB of L2 cache.

\subsection{Evaluation Metrics} 
To ensure a robust assessment of local tracking accuracy, we employ the frame-to-frame relative transform error, following standard evaluation methodologies \cite{sturm2012benchmark}. This metric is less sensitive to long-term drift than absolute trajectory error. For each consecutive frame pair $(i, i+1)$, we compute the error between the estimated and ground truth relative poses. The estimated relative pose is defined as:
\begin{align}
\mathbf{T}_{rel}^{est} = (\mathbf{T}_i^{est})^{-1} \mathbf{T}_{i+1}^{est}
\end{align}
while the ground truth relative pose is:
\begin{align}
\mathbf{T}_{rel}^{gt} = (\mathbf{T}_i^{gt})^{-1} \mathbf{T}_{i+1}^{gt}
\end{align}
The relative pose error is then decomposed into two components:

\paragraph{Translation Error ($e_{trans}$)} 
The Euclidean distance between the relative translation vectors:
\begin{align}
e_{trans} = ||\mathbf{t}_{rel}^{est} - \mathbf{t}_{rel}^{gt}||_2
\end{align}

\paragraph{Rotation Error ($e_{rot}$)} 
The angular difference between the relative rotation matrices:
\begin{align}
e_{rot} = \arccos\left(\frac{\text{trace}((\mathbf{R}_{rel}^{est})^T \mathbf{R}_{rel}^{gt}) - 1}{2}\right)
\end{align}

The results presented in Table~\ref{tab:euroc_combined} are the Root Mean Square Error (RMSE) of these metrics over all frame pairs in each sequence. 
\subsection{Quantitative Performance Analysis} 
As summarized in Figure~\ref{fig:comparison} and Table~\ref{tab:euroc_combined}, our experiments demonstrate a progressive improvement in tracking accuracy with each phase, where the final learning-based approach in \textbf{Phase 2} achieves the highest performance. Overall, Phase 2 attains the lowest error in the majority of test cases, specifically in 10 out of 11 sequences for both translation and rotation metrics. A notable exception was observed in the highly challenging \texttt{V2\_03} sequence, where rapid motion and challenging conditions seemed to highlight the complementary strengths of different approaches. Despite this outlier, the results support the overall effectiveness of our proposed framework.

The performance gains align with the intended effects of our two-phase approach. The initial comparison of Phase 1 to the Baseline showed modest improvements of 3--11\% in most sequences, suggesting that basic uncertainty modeling may provide benefits over uniform weighting assumptions. This lends support to the fundamental idea of considering measurement reliability in VIO systems. Building on this foundation, Phase 2 demonstrated more substantial additional improvements of 4--25\% over Phase 1, with the larger gains typically observed in sequences with richer geometric structure. This outcome is consistent with the hypothesis that statistical learning can capture complex uncertainty patterns that simple geometric models may not fully represent.

Cumulatively, the complete framework in Phase 2 resulted in overall accuracy improvements ranging from 8--34\% compared to the Baseline across different sequences, suggesting clear benefits from the statistical uncertainty learning process. The degree of improvement appeared to correlate with sequence characteristics and motion complexity. On sequences with stable motion and good visual features (e.g., \texttt{V1\_01}, \texttt{MH\_01}), Phase 2 achieved substantial error reductions of approximately 25--34\%. This trend extended to sequences with more complex motion patterns (e.g., \texttt{V1\_02}, \texttt{V1\_03}, \texttt{MH\_03}), where the adaptive nature of Phase 2 continued to yield meaningful gains in the range of 15--25\%.

Interestingly, the most challenging sequence, \texttt{V2\_03}, presented a mixed result that may highlight fundamental trade-offs under extreme conditions. In this aggressive scenario, no single method excelled consistently across both metrics; the Baseline achieved the most stable translation performance, while Phase 1 with geometric constraints produced the best rotation accuracy.
\subsection{Computational Efficiency} These accuracy improvements were observed with what appears to be minimal computational overhead. The Phase 1 uncertainty propagation added approximately a 2.3\% computational cost, while the learning-based calibration in Phase 2 added a mere 0.8\% on top of Phase 1. On our experimental system, equipped with an Intel Core i7-14700KF CPU, the entire framework maintained real-time performance at {90--100 Hz}, suggesting its potential suitability for practical visual-inertial odometry applications.

\section{Conclusion}

Our findings suggest that the explicit modeling of measurement uncertainty is a valuable component of visual-inertial odometry. The initial phase of our study indicated that incorporating even a basic geometric uncertainty model can yield modest performance gains of 3-11\% over assuming uniform covariance. This observation highlights the potential importance of properly weighting observations rather than treating them as equally reliable. Building on this, the second phase showed that a statistical learning approach could offer further improvements of 4-25\% over the geometric model. This suggests that data-driven methods may be capable of capturing more complex uncertainty patterns that are not fully represented by simpler, pre-defined models.

Taken together, the progressive comparison from our baseline to Phase 1 and Phase 2 resulted in overall trajectory accuracy improvements ranging from 8-34\% across different sequences. This progression lends support to the value of both foundational uncertainty modeling and more advanced statistical learning techniques. Furthermore, the proposed framework demonstrated practical viability, maintaining real-time performance at 90-100 Hz with a computational overhead of less than 3\%, suggesting its suitability for autonomous navigation tasks.

In summary, the systematic comparison presented in Table~\ref{tab:euroc_combined} is consistent with our core hypothesis that principled uncertainty modeling is beneficial and that statistical learning represents a promising direction for enhancing traditional geometric approaches in visual-inertial state estimation. Future work will investigate extending this framework to dynamic environments, exploring uncertainty learning for semantic features, and analyzing theoretical convergence properties of the learning algorithm. Additionally, we plan to develop more robust approaches that can maintain superior stability and accuracy in highly challenging scenarios such as the \texttt{V2\_03} sequence, where rapid motion and extreme conditions currently present limitations for statistical learning methods.

\bibliography{references}
\end{document}